\pdfoutput=1

\documentclass[11pt]{article}

\usepackage{acl}

\usepackage{times}
\usepackage{latexsym}

\usepackage[T1]{fontenc}

\usepackage[utf8]{inputenc}

\usepackage{microtype}

\usepackage{inconsolata}

\usepackage{algorithm}
\usepackage{algpseudocode}
\usepackage{booktabs}       
\usepackage{amsfonts}       
\usepackage{nicefrac}       
\usepackage{microtype}      
\usepackage{adjustbox}
\usepackage{tabularx}
\usepackage{array}
\usepackage{amsmath}
\usepackage{amsthm}
\usepackage{blindtext}
\usepackage{mathtools}
\usepackage{enumitem}
\usepackage{multirow}
\usepackage{booktabs}
\usepackage{graphicx}
\usepackage{bm}
\usepackage{threeparttable}
\usepackage{float}

\theoremstyle{definition}

\newtheorem{definition}{Definition}

\newtheorem{proposition}{Proposition}

\title{Differentially Private Natural Language Models: Recent Advances and Future Directions}

\author{
Lijie Hu$^1$,
Ivan Habernal$^2$,
Lei Shen$^3$,
and Di Wang$^1$ \\
$^1$CEMSE, King Abdullah University of Science and Technology\\
$^2$TrustHLT, Technical University of Darmstadt\\
$^3$JD AI Research, Beijing, China\\
\{lijie.hu, di.wang\}@kaust.edu.sa,
ivan.habernal@tu-darmstadt.de,
shenlei20@jd.com
}

\begin{document}
\maketitle
\begin{abstract}
Recent developments in deep learning have led to great success in various natural language processing (NLP) tasks. However, these applications may involve data that contain sensitive information. Therefore, how to achieve good performance while also protecting the privacy of sensitive data is a crucial challenge in NLP. To preserve privacy, Differential Privacy (DP), which can prevent reconstruction attacks and protect against potential side knowledge, is becoming a de facto technique for private data analysis. In recent years, NLP in DP models (DP-NLP) has been studied from different perspectives, which deserves a comprehensive review. In this paper, we provide the first systematic review of recent advances in DP deep learning models in NLP. In particular, we first discuss some differences and additional challenges of DP-NLP compared with the standard DP deep learning. Then, we investigate some existing work on DP-NLP and
present its recent developments from three aspects: gradient perturbation based methods, embedding vector perturbation based methods, and ensemble model based methods. We also discuss some challenges and future directions.
\end{abstract}

\section{Introduction}
The recent advances in deep neural networks have led to significant success in various tasks in Natural Language Processing (NLP), such as sentiment analysis, question answering, information retrieval, and text generation. However, such applications always involve data that contains sensitive information. For example, a model of aid typing on a keyboard which trained from language data might contain sensitive information such as passwords, text messages, and search queries. Moreover, language data can also identify a speaker explicitly by name or implicitly, for example, via a rare or unique phrase. 
Thus, one often encountered challenge in NLP is how to handle this sensitive information. To overcome the challenge, 
privacy-preserving NLP has been intensively studied in recent years. One of the commonly used approaches is based on text anonymization \citep{pilan2022text}, which identifies sensitive attributes and then replaces these sensitive words with some other values. Another approach is injecting additional words into the original text without detecting sensitive entities in order to achieve text redaction \citep{sanchez2016c}. However, removing personally identifiable information or injecting additional words is often unsatisfactory, as it has been shown that an adversary can still infer an individual’s membership in the dataset with high probability via the summary statistics on the datasets \citep{narayanan2008robust}. Moreover, recent studies claim that deep neural networks for NLP tasks often tend to memorize their training data, which makes them vulnerable to leaking information about training data \citep{shokri2017membership,carlini2021extracting,carlini2019secret}. One way that takes into account the limitations of existing approaches by preventing individual re-identification and protecting against any potential data reconstruction and side-knowledge attacks is designing  Differentially Private (DP) algorithms. DP \citep{dwork2006calibrating} provides provable protection against identification and is resilient to arbitrary auxiliary information that might be available to attackers. Thanks to its formal guarantees, DP has become a de facto standard tool for private statistical data analysis. 

Although there are numerous studies on DP machine learning and DP deep learning, such as \citep{abadi2016deep,bu2020deep,yu2019differentially}, most of them mainly focus on either the continuous tabular data or image data, and less attention has been paid to adapting variants of DP algorithms to the context of NLP and the text domain. On the other side, while there are several surveys on DP and its applications, such as \citep{ji2014differential,dankar2013practicing,DBLP:journals/scn/XiongLLCN20,s20247030,Desfon+2020+288+313}, none of them study its applications to the NLP domain. Recently, \citet{klymenko2022differential} gave a brief introduction to applications of DP in NLP, but the reviewed work is not exhaustive, and it lacks a technical and systematic view of DP-NLP. Thus, to fill in this gap, in this paper, we provide the first technical overview of the recent developments and challenges of DP in language models.

Specifically, we give a survey on the most recent 70\footnote{Note that we did not cover all related works, see the Limitations and Future Directions sections for the works that are not included in this paper.} papers on deep learning based approaches for NLP tasks under DP constraints. First, we show some specificities of DP-NLP  compared with the general deep learning with DP. Then we discuss current results from three perspectives via the ways of adding randomness to ensure DP: the first one is gradient perturbation based methods which includes DP-SGD and DP-Adam; the second one is embedding vector perturbation based methods which includes DP auto-encoder; the last one is ensemble model based methods which includes PATE. For each type of approach, we also consider its applications to different NLP tasks. Finally, we present some potential challenges and future directions.

Due to space limits, in Appendix \ref{sec:dp}, we give a preliminary introduction to DP to readers who are unfamiliar with DP. 

\begin{table*}[!htbp]
\vspace{-2.0em}
\centering
\resizebox{1.0\textwidth}{!}{
\begin{tabular}{ccccccc}
\toprule
\textbf{Method Type} & \textbf{Publications} & \textbf{Scenarios} & \textbf{Definition} & \textbf{Model Architecture} & \textbf{DP Level} & \textbf{Downsteam Tasks}  \\
\midrule
\multirow{35}{*}{\begin{tabular}[c]{@{}c@{}}\textbf{Gradient} \\ \textbf{Perturbation}\\\textbf{Based} \\ \textbf{Methods}\end{tabular}}          
& \citet{hoory-etal-2021-learning-evaluating}            &   \multirow{4}{*}{\begin{tabular}[c]{@{}c@{}} \textbf{Pre-trained}\end{tabular}}       & \multirow{2}{*}{\begin{tabular}[c]{@{}c@{}} DP\end{tabular}}          & BERT                  & Sample-level        & Entity-extraction      \\
      & \citet{anil2021large}      &      &        &  BERT                  & Sample-level        & ---      \\
            & \citet{yin2022privacy}     &      &        &  BERT                  & Sample-level        & Classification, QA    \\
                        & \citet{senge2022one}     &      &        &  BERT, XtremeDistil                 & Sample-level        & Classification, NER, POS, QA    \\
& \citet{ponomareva2022training}  &      &        &  T5                 & Sample-level        & NLU \\ 
                
\cmidrule{2-7}
    & \citet{Finetuning2022Yu}            &   \multirow{20}{*}{\begin{tabular}[c]{@{}c@{}} \textbf{Fine-tuning}\end{tabular}}       & \multirow{20}{*}{\begin{tabular}[c]{@{}c@{}} DP\end{tabular}}          & RoBERT, GPT-2                  & Sample-level        & NLG, NLU      \\
    & \citet{Repara2021Yu}            &          &           & BERT                  & Sample-level        & Classification, NLU      \\
    & \citet{dupuy2022efficient}            &          &           & BERT,BiLSTM                  & Sample-level        & Classification, NER  
    \\
    & \citet{li2021large}            &          &           & GPT-2, (Ro)BERT                  & Sample-level        & Classification, NLG     \\
  & \citet{lee2023private}    &          &           & GPT-2, DialoGPT               & Sample-level        & Meeting Summarization  \\ 
& \citet{xia2023differentially}  &          &           & GPT-2, (Ro)BERT                  & Sample-level        & Classification    \\
& \citet{behnia2022ew}   &          &           &  (Ro)BERT                  & Sample-level        & NLU    \\
& \citet{bu2023differentially} &          &           & GPT-2, (Ro)BERT                  & Sample-level        & Classification    \\
& \citet{gupta2023jointly}&          &           & (Ro)BERT                  & Sample-level        & GLU   \\ 
  &\citet{du2021dp}   &          &          & GPT-2, (Ro)BERT                  & Sample-level        & Classification, NLG    \\  
& \citet{bu2022differentially}  &          &           & (Ro)BERT                  & Sample-level        & Classification, NLG  \\  
& \citet{yue2022synthetic}   &          &           & GPT-2                  & Sample-level        & Synthetic Text Generation  \\  
& \citet{mireshghallah2022privacy}  &          &           & GPT-2                  & Sample-level        & Synthetic Text Generation  \\  
&\citet{carranza2023privacy}   &          &           & T5                & Sample-level        & Query Generation   \\  
    & \citet{Timour2021GCN}            &          &           & GPT-2                 & Sample-level        & Classification      \\
       & \citet{al2021differentially}   & &  & GPT-2                 &  Sample-level        & Synthetic Text Generation  \\ 
          & \citet{wunderlich2021privacy}       &    &   &     BERT,CNN    &   Sample-level      &  Classification     \\  
        & \citet{li2022private}       &    &   &     LSTM   &   Sample-level      &  Classification     \\  
        & \citet{amid2022public}  &    &   &     LSTM   &   Sample-level      &  Classification     \\   
              & \citet{shi2021selective}            &          &   \textbf{SDP}        & RNN             & Sample-level        &  NLG, Dialog System    \\
            & \citet{shi2022just}       &          &   \textbf{SDP}        &     GPT-2, (Ro)BERT    & Sample-level        &  NLG, NLU   \\  
\cmidrule{2-7}
    & \citet{mcmahan2017learning}            &         &        & LSTM, RNN                  & User-level        & Prediction, Classification      \\
& \citet{ramaswamy2020training}    &         &        & LSTM                 & User-level        & Prediction, Classification      \\ 
& \citet{kairouz2021practical}  &         &       & LSTM              & User-level, Sample-level        & Prediction, Classification      \\ 
& \citet{choquette2022multi}  &   \textbf{Federated Learning}    &    \textbf{LDP}     & LSTM              & User-level, Sample-level        & Prediction    \\  
 & \citet{koloskova2023convergence}  &        &        & LSTM              & User-level, Sample-level        & Prediction    \\ 
 & \citet{denisov2022improved} &        &       & LSTM              & User-level, Sample-level        & Prediction  \\
 &\citet{agarwal2021skellam} &        &       & LSTM              & User-level, Sample-level        & Prediction  \\ 
 & \citet{wang2023can} &        &       & LaMDA             & User-level      & Prediction  \\  
 & \citet{xu2023federated} &        &       & Gboard            & User-level      & Prediction  \\  
\midrule
\midrule
\multirow{28}{*}{\begin{tabular}[c]{@{}c@{}}\textbf{Embedding} \\ \textbf{Vector} \\ \textbf{Perturbation}\\\textbf{Based} \\ \textbf{Methods}\end{tabular}} 
      & \citet{lyu2020towards}            & \multirow{7}{*}{\begin{tabular}[c]{@{}c@{}} Private Embedding  \end{tabular}}  &    \multirow{7}{*}{\begin{tabular}[c]{@{}c@{}} \textbf{LDP} \end{tabular}}       &   BERT                &  Word-level       & Classification      \\
      & \citet{lyu-etal-2020-differentially}            &   &          & BERT                  & Word-level        & Classification      \\
      & \citet{plant-etal-2021-cape}            &   &           &       BERT            &   Word-level      &  Classification     \\
      & \citet{krishna-etal-2021-adept}           &   &           &     Auto-Encoder             &   Word-level      &  Classification     \\
      & \citet{habernal-2021-differential}           &   &          &             Auto-Encoder     &   Word-level      &   Classification    \\
  & \citet{DBLP:conf/sbp-brims/AlnasserB021}    &   &          &            Auto-Encoder    &   Word-level      &   Classification  \\
      & \citet{Igamberdiev.2022.COLING} &   &          &             Auto-Encoder     &   Word-level      &   Classification    \\
      & \citet{maheshwari2022fair}            &  &
      &   Auto-Encoder  &   Word-level      &   Classification \\
      &\citet{DBLP:journals/corr/abs-2309-10551}    &  &
      &  GloVe  &   Word-level      &   Classification \\
      & \citet{DBLP:conf/acl/ChenMWCN0C23}   &  &
      &  GloVe, BERT  &   Token-level      &   Classification \\ 
          & \citet{DBLP:journals/corr/abs-2309-06746} &  Fine-tuning           & Sequence LDP & BERT & Sentence-level & Classification, QA      \\ 
\cmidrule{2-7}
\cmidrule{3-3}
      & \citet{meehan2022sentence} &  \multirow{1}{*}{\begin{tabular}[c]{@{}c@{}} Private Embedding \end{tabular}} &     \multirow{1}{*}{\begin{tabular}[c]{@{}c@{}} DP \end{tabular}}      &    SBERT              &   \textbf{Sentence-level}      &  Classification     \\
\cmidrule{2-7}
     & \citet{feyisetan2020privacy}           & \multirow{9}{*}{\begin{tabular}[c]{@{}c@{}} Private Embedding \end{tabular}}  &    \multirow{10}{*}{\begin{tabular}[c]{@{}c@{}} \textbf{LMDP} \end{tabular}}       &    GloVe, BiLSTM       &   Word-level      &  Classification, QA     \\
     & \citet{xu2020mahala}           &   &           &      GloVe             &  Word-level   &   Classification    \\
     & \citet{xu2021vickrey}           &   &           &     GloVe,FastText              &  Word-level       & Classification       \\
     & \citet{xu2021density}           &   &           &      GloVe, CNN             &     Word-level    &   Classification     \\
     & \citet{carvalho2021tem}           &   &           &      GloVe             &  Word-level       & Classification      \\
     & \citet{feyisetan2021private}           &   &            &       GloVe, FastText   &   Word-level      &  Classification  \\
    & \citet{feyisetan2019leveraging}           &   &           &       GloVe           &  Word-level       &   Classification, Prediction    \\
    & \citet{carvalho2021brr}           &   &           &       GloVe, FastText      &   Word-level      & Classification       \\
    & \citet{tang2020privacy}           &   &           &       GloVe            &    Word-level     &   Classification    \\
    & \citet{DBLP:conf/uai/ImolaKWAT22}   &   &           &       GloVe, FastText      &   Word-level      & Classification    \\
    & \citet{DBLP:journals/corr/abs-2306-01457} &   &           &       GloVe      &   Word-level      & Classification    \\ 
    & \citet{DBLP:journals/corr/abs-2306-01471} &   &           &       GloVe      &   Word-level      & Classification    \\  
    & \citet{qu2021natural}      &  Fine-tuning   &           & BERT, BiLSTM                  & Token-level        & Classification,NLU      \\
    & \citet{10.1145/3543507.3583512} & Private Embedding & & BERT & Sentence-level & Classification, QA \\
    & \citet{DBLP:journals/corr/abs-2305-06212}  & Private Prompt Tuning & & BERT, TA & Word-level & Classification, QA \\ 
\cmidrule{2-7}
    & \citet{yue-etal-2021-differential}           & Private Embedding  &    \textbf{UMLDP}       &    BERT, GloVe      &   Word-level      &   Classification,QA    \\
        \midrule
    \midrule
\multirow{4}{*}{\begin{tabular}[c]{@{}c@{}}\textbf{Ensemble} \\ \textbf{Model} \\ \textbf{Based} \\ \textbf{Methods}\end{tabular}} 
      & \citet{DBLP:journals/corr/abs-2305-15594}            & \multirow{3}{*}{\begin{tabular}[c]{@{}c@{}} \textbf{In-context Learning} \end{tabular}}  &      &  GPT-3              &    Sample-level   &   Classification   \\
      & \citet{wu2023privacypreserving}            &   &          & GPT-3                  & Sample-level        & Classification, QA, Dialog Summarization     \\
      & \citet{DBLP:journals/corr/abs-2309-11765}           &   &           &       GPT-3          &   Sample-level      & Classification, Information Extraction    \\
      & \citet{DBLP:conf/nips/TianZHWZH22}      & \textbf{Fine-tuning}  &           &     GPT-2             &   Sample-level, User-level      &  Synthetic Text Generation     \\

\bottomrule                  
\end{tabular}
}
\caption{ An overview of studies for DP-NLP.\label{overview}}
\vspace{-7pt}
\end{table*}

\section{Specificities of NLP with DP}
We first discuss some specificities for DP-NLP compared with the standard DP deep learning. Generally speaking, there are two aspects: one is privacy notations, and another is privacy levels.

\subsection{Variants of DP Notions in NLP}
Recall that DP ensures data analysts or adversaries will get almost the same information if we change any single data sample in the training data, i.e., it treats all records as sensitive.  However, such an assumption is quite stringent. On the one side, unlike image data, for text data, it is more common that only several instead of all attributes need to be protected.  For example, for the sentence "My cell phone number is 1234567890", only the last token with the actual cell phone number needs to be protected. On the other side, canonical DP requires that the log of the ratio between the distribution probabilities is always upper bounded by the privacy parameter $\epsilon$ for any pair of neighboring data. However, such a requirement is also quite restrictive. For example, for the sentence "I will arrive at 2:00 pm", we want the adversary not to distinguish it from the sentence "I will arrive at 4:00 pm". However, DP also can ensure the adversary cannot distinguish it from the sentence "I will arrive at 100:00 pm", which is meaningless. Thus, for language data, besides the canonical DP, it is also reasonable to study its relaxations for some specific scenarios. Actually, this is quite different from the existing work on DP deep learning, which mainly focuses on standard DP definitions. In the following, we will discuss some commonly used relaxations of DP for language models. 

\paragraph{SDP.} As we mentioned above, in some scenarios, the sensitive information in text data is sparse, and we only need to protect some sensitive attributes instead of the whole sentence. Based on this, \citet{shi2021selective} propose a new privacy notion, namely selective differential privacy (SDP), to provide privacy guarantees on the sensitive portion of the data to improve model utility. From the definition aspect, the main difference between SDP and DP is the definition of neighboring datasets. Informally, in SDP, two datasets are adjacent if they differ in at least one 
sensitive attribute. However, it is hard to define such neighboring datasets directly as there are some correlations between sensitive and non-sensitive attributes, indicating that we can still infer information on sensitive attributes \cite{DBLP:conf/sigmod/KiferM11}. To address the issue,  \citet{shi2021selective}  leverage the Pufferfish framework in \cite{DBLP:journals/tods/KiferM14}. 

\paragraph{Metric DP.} To relax the requirement that the log probability ratio is uniformly bounded by $\epsilon$ for all neighboring data pairs,  \citet{feyisetan2020privacy} first adopt the Metric DP (or $d_\chi$-privacy) to the problem of private embedding, which is proposed by \cite{chatzikokolakis2013broadening} for location data originally. In particular, a Metric DP mechanism could report a token in a privacy-preserving manner while giving a higher probability to tokens that are close to the current token, and a negligible probability to tokens in a completely different part of the vocabulary, where we will use some distance function $d$ to measure the distance between two tokens. 
\begin{definition}
For a data domain (vocabulary) $\mathcal{X}$, a randomized algorithm $\mathcal{A}:\mathcal{X}\mapsto \mathcal{R}$ is called $(\varepsilon, \delta)$-Metric DP with distance function $d$ if 
for any $S, S' \in \mathcal{X}^l$ and $T\subseteq \mathcal{R}$ we have 
\begin{equation*}
   {\text{Pr}[\mathcal{A}(S)\in T] }\leq e^{d(S, S')\varepsilon} \text{Pr}[\mathcal{A}(S')\in T]+\delta.
\end{equation*}
\end{definition}

From the above definition, we can see the probability ratio of observing any particular output $y$ given two possible inputs $S$ and $S'$ 
is bounded by $e^{\varepsilon d(S',S)}$ instead of $e^\epsilon$ in DP. Motivated by Metric DP and local DP, \cite{feyisetan2020privacy} provides the Local Metric DP (LMDP) and uses it for private word embeddings  (see Section \ref{sec:embedding} for details). Motivated by Utility-optimized LDP (ULDP) \cite{DBLP:conf/uss/Murakami019} rather than LDP,  recently \citet{yue-etal-2021-differential} propose Utility-optimized Metric LDP (UMLDP). It exploits the fact that different inputs have different sensitivity levels to achieve higher utility. By assuming the input space, such as the set of tokens is split into sensitive and non-sensitive parts, UMLDP achieves a privacy guarantee equivalent to LDP for sensitive inputs. 

\subsection{Variants Levels of Privacy in NLP} 
When we consider using DP, the first question is what kind of information we aim to protect. In the previous studies on DP deep learning, we always wanted to protect the whole data sample. However, in the NLP domain, such one data sample could be either a word, a sentence, a paragraph, etc. If we ignore the concrete privacy level and directly apply the previous DP  methods, we may have mediocre results. Thus, unlike the sample level privacy in DP deep learning, researchers in NLP consider different levels of privacy. Especially, they focus on the word level and sentence level, which aims to protect each word and sentence respectively \cite{meehan2022sentence,feyisetan2019leveraging}. 

In the federated learning setting, there is a central server and several users each of them has a local dataset, the sample level of DP may be insufficient. For example, in language modeling, each user may contribute many thousands of words to the training data, and each typed word makes its own contribution to the RNN’s training objective. In this case, just protecting each word is unsatisfactory, and it is still possible to re-identify users. Thus, besides the sample level, we also have the user level of privacy, which aims to protect users' histories.  
After discussing some specificities of DP-NLP. In the following, we categorize its recent studies into three classes based on their methods to ensure DP: gradient perturbation based methods, embedding vector perturbation based methods, and ensemble model based methods. See Tab. \ref{overview} for an overview.  

\section{Gradient Perturbation Based Methods}
Given a training dataset with $n$ samples $D=\{x_i\}_{i=1}^n$, a loss function (such as cross-entropy loss) is defined to train the model, which takes the parameter $\theta\in \mathbb{R}^d$ of neural network and samples and outputs a real value: 
\begin{equation}\label{eq:2}
    L(\theta, D)= \sum_{i=1}^n \ell(\theta, x_i).
\end{equation}  
The goal is to find the weights of the network that minimizes $ L(\theta, D)$, {\em i.e.,} $\theta^*= \arg\min_{\theta} L(\theta, D)$. With additional constraint on DP, now we aim to design an $(\varepsilon, \delta)/\varepsilon$-DP  algorithm $\mathcal{A}$ to make the private estimated parameter $\theta_{priv}$ close to $\theta^*$. 

\noindent {\bf Example:} In Language Modeling (LM), we have a corpus $D=\{x_1, \cdots, x_n\}$ where each text sequence $x_i$ consists of multiple tokens $x_i=(x_{i1}, \cdots, x_{im_i})$ with $x_{ij}$ as the $j$-th token of $x_i$. The
goal of LM is to train a neural network (e.g., RNN) parameterized by $\theta$ to learn the probability of the sequence $p_\theta(x)$, which can be represented as the following objective function
\begin{align*}
-\sum_{i=1}^n \sum_{j=1}^{m_i} \log p_\theta(x_{ij}|x_{i1},\cdots, x_{i(j-1)}).
\end{align*}

We first review the DP-SGD method \cite{abadi2016deep}. In the non-private case, to minimize the objective function (\ref{eq:2}), the most fundamental method is SGD, i.e., in the $t$-th iteration, we update the model as follows:
\begin{equation*}
    \theta^{t+1}=\theta^t-\eta \frac{1}{|B|
}\sum_{x\in B} \nabla \ell(\theta^t, x), 
\end{equation*}
where $B$ is a subsampled batch of random examples, $\eta$ is the learning rate and $\theta^t$ is the current parameter. DP-SGD modifies the SGD-based methods by adding Gaussian noise to perturb the (stochastic) gradient in each iteration of the training, {\em i.e.,}  during the $t$-th iteration DP-SGD will compute a noisy gradient as follows: 
\begin{equation}\label{equ:dpsgd}    
g^t = \frac{1}{|B|}(\sum_{x_i \in B}\limits \hat{g}^t_{i} +\mathcal{N}\left(0, \sigma^2 C^2 I_d\right)), 
\end{equation}
$\sigma$ is the noise multiplier, $\hat{g}^t_{i}$ is some vector computed from $\nabla \ell(\theta^t, x_i)$ and $g^t$ is the (noisy) gradient used to update the model. The main reason here we use $\hat{g}^t_{i}$ instead of the original gradient vector is that we wish to make the term $\sum\limits \hat{g}^t_{i}$ has bounded $\ell_2$-sensitivity so that we can use the Gaussian mechanism to ensure DP. The most commonly used approach to get a $\hat{g}^t_{i}$ is clipping the gradient: $\hat{g}^t_{i}=\nabla \ell(\theta^t, x_i)\min \{1, \frac{C}{\|\nabla \ell(\theta^t, x_i)\|_2}\}$ {\em i.e.}, each gradient vector is clipped by a hyper-parameter $C>0$. Since the $\ell_2$-sensitivity of $\sum\limits \hat{g}^k_{i}$ is bounded by $C$, after the clipping, we can add Gaussian noise to ensure DP. As there are several iterations, and in each iteration, we use some subsampling strategy, we can use the composition theorem and privacy amplification to compute the total privacy cost of DP-SGD. Equivalently, given a fixed privacy budget $(\epsilon, \delta)$, number of iterations and subsampling strategy, one can get the minimal noise multiplier $\sigma$ to ensure DP, see \cite{DBLP:journals/jsait/AsoodehLCKS21, DBLP:conf/nips/GopiLW21, DBLP:journals/corr/abs-1908-10530, DBLP:journals/jpc/WangBK20, DBLP:conf/icml/ZhengDLS20, DBLP:conf/icml/ZhuW19} for details.

Its main idea is to use the noisy and clipped subsampled gradient $g^t$ to approximate the whole gradient $\nabla L(\theta^t, D)$. In fact, besides SGD, we can use this idea for any optimizer, such as Adam \cite{DBLP:journals/corr/KingmaB14}, whose private version DP-Adam is proposed and applied in BERT by \cite{anil2021large}. In the past few years, there has been a long list of work on DP-SGD from different perspectives, such as the subsampling strategy, faster clipping procedures, private clipping parameter tuning, and the selection of batch size. In the following, we will only discuss the previous work on using DP-SGD-based methods for variants of NLP tasks.

\subsection{DP Pre-trained Models}
Recent developments in NLP have led to successful applications in large-scale language models with the appearance of transformer \citep{devlin2018bert}. It combines the contextual information into language models with a more powerful ability of representation. These models are called pre-trained models, which train word embedding in large corpora targeting various tasks and gain the knowledge for downstream tasks \citep{peters-etal-2018-deep}. In this section, we review some papers that focus on pre-trained NLP models under DP constraints. 

The workflow of BERT \citep{devlin2018bert} is pre-training the unlabeled text using some large corpora first. Then, the downstream tasks first initialize the model using the same parameters and fine-tune the parameters according to different tasks. Despite the benefits of powerful representation ability given by the pre-training process, it also has privacy issues since the model would memorize sensitive information such as words or phrases.
 
In order to solve this privacy leakage issue, there are several studies on how to train BERT privately. \citet{hoory-etal-2021-learning-evaluating} successfully trained a differentially private BERT model by modifying the WordPiece algorithm to satisfy DP, and conducted experiments on the problem of entity extraction tasks from medical text. They construct a tailored domain-specific DP-based trained vocabulary designed to generate a new
domain-specific vocabulary while maintaining user privacy and then use the original DP-SGD in the training process. For the DP vocabulary part, they first construct a word histogram by dividing the text into a sequence of $N$-word tuples and then add Gaussian noise to the histogram to ensure $(\epsilon, \delta)$-DP. Finally, they clip the histogram with some threshold. For the training phase, they use the original DP-SGD to meet privacy guarantees. Besides, they also use the parallel training trick to make the training faster. Very recently, \citet{yin2022privacy} applied DP-BERT to the legal NLP domain. While DP-BERT can achieve good performance with privacy guarantees in language tasks. There are still two problems: a large gap between non-private accuracy and private accuracy, and computation inefficiency of clipping every sample gradient in DP-SGD. In order to mitigate these issues, \citet{anil2021large} later privatizes the Adam optimizer to improve the performance. Instead of adding noise and clipping every entry in every batch in DP-SGD, it selects a pre-defined number of samples randomly and sums the clipped gradients of these selected samples, then it updates average gradients with Gaussian noise adding the sum in each batch. Besides, it also uses an increasing batch size schedule instead of a fixed one. It finds that large batch size can improve accuracy, and the increasing batch size schedule can improve training efficiency. \cite{senge2022one} recently studied five different typical NLP tasks with varying complexity using modern neural models based on BERT and XtremeDistil architectures. They showed that to achieve adequate performance, each task and privacy regime requires special treatment. 

Besides BERT, \citet{ponomareva2022training} privately pre-train T5 \cite{raffel2020exploring} via their proposed private tokenizer called DP-SentencePiece and DP-SGD. They show that DP-T5 does not suffer a large drop in pre-training utility, nor in training speed, and can still be fine-tuned to high accuracy on downstream tasks.
  
\subsection{DP Fine-tuning}
Besides training pre-trained models using DP algorithms, another direction is how to fine-tune pre-trained models privately. Here, the main difference is that we assume the pre-trained models, such as BERT have been trained with some public data, and our goal is to privately fine-tune targeting specific downstream tasks that involve sensitive data. It is noted that in this section, we also include some related work on training shallow neural networks in DP such as RNN or LSTM such as \cite{li2022private,amid2022public} as these methods can be directly applied to DP fine-tuning. 

In this topic, the first direction is to investigate different tasks in the DP model and to compare its performance compared to the non-private one for studying the utility-privacy trade-off. \citet{yue2022synthetic} consider the task of synthetic text generation and show that simply fine-tuning a pre-trained GPT-2 with the vanilla DP-SGD enables the model to generate useful synthetic text. \citet{mireshghallah2022privacy} recently extended to generating latent semantic parses in the DP model and then generating utterances based on the parses. \citet{carranza2023privacy} use DP-SGD to fine-tune a publicly pre-trained LLM on a query generation task. The resulting model can generate private synthetic queries representative of the original queries which can be freely shared for downstream non-private recommendation training procedures. Very recently, \citet{lee2023private} adopted the DP-SGD to the meeting summarization task and showed that DP can improve performance when evaluated on unseen meeting types. \citet{al2021differentially} use GPT-2 and DP-SGD based methods to generate synthetic EHR data which can de-identify sensitive information for clinical text. \citet{wunderlich2021privacy} study the hierarchical text classification task, and they use DP-SGD to Bag of Words (BoW), CNNs and Transformer-based architectures. They find that Transformer-based models achieve better performance than CNN-based models in large datasets, while CNN-based models are superior to Transformer-based models in small datasets. 

The second direction is to reduce the huge memory cost of storing individual gradients and decrease the added noise, which suffers notorious dimensional dependence in DP-SGD. Specifically, the studies in this direction always propose a general method for DP-SGD and then perform the method for different NLP tasks. 
\citet{Repara2021Yu} propose a variant of DP-SGD called the Reparametrized Gradient Perturbation (RGP) method.  The framework of RGP parametrizes each weight matrix with two low-rank carrier matrices and a residual weight matrix, which will be used to approximate the original one. Such a way can reduce the memory cost for computing individual gradient matrices and can maintain the optimization process via forward/backward signals. Later, based on RGP, \citet{Finetuning2022Yu} show that advanced parameter-efficient methods such as \cite{houlsby2019parameter,karimi2021compacter} can lead to simpler and
significantly improved algorithms for private fine-tuning. Instead of DP-SGD, \citet{du2021dp} propose a DP version of Forward-Propagation. Specifically, it clips representations followed by noise addition in the forward propagation stage. 

Besides adapting the optimization method in vanilla DP-SGD, there are also some works on modifying the clipping operation or the fine-tuning method directly to save the memory cost. \citet{li2021large} propose a memory-saving technique that allows clipping in DP-SGD for fine-tuning to run without instantiating per-example gradients for any linear layer in the model. The technique enables private training Transformers with almost the same memory cost as non-private training at a modest run-time overhead. \citet{dupuy2022efficient} propose another variant of DP-SGD via micro-batch computations per GPU and noise decay and apply it to fine-tuning models. Specifically, they scale gradients in each micro-batch and set a decreasing noise multiplier with epoch. Then, they add scaled Gaussian noise to gradients. In this way, they can make the training faster and adapt it for GPU training. \citet{bu2023differentially} develop a novel Book-Keeping (BK) technique that implements existing DP optimizers, with a substantial improvement on the computational cost while also keeping almost the same accuracy as DP-SGD. \citet{gupta2023jointly} propose a novel language transformer finetuning strategy that introduces task-specific parameters in multiple transformer layers. They show that the method of combining RGP and their novel strategy is more suitable for low-resource applications. \citet{bu2022differentially} privatize the bias-term fine-tuning (BiTFiT) and show that DP-BiTFiT matches the state-of-the-art accuracy for DP algorithms and the efficiency of the standard BiTFiT \cite{zaken2022bitfit}. \citet{Timour2021GCN} apply DP-Adam in Graph Convolutional Networks to perform the private fine-tuning for text classification. Specifically, they first split the graph into disconnected sub-graphs and then add noise to gradients.

Rather than reducing the memory cost, there are some papers considering developing variants of the DP-SGD method to improve performance. For example, \citet{xia2023differentially} propose a per-sample adaptive clipping algorithm, which is a new perspective and orthogonal to dynamic adaptive noise and coordinate clipping methods. \citet{behnia2022ew} use the Edgeworth accountant \cite{wang2022analytical} to compute the amount of noise that is required to be added to the gradients in SGD to guarantee a certain privacy budget, which is lower than the original DP-SGD. \citet{li2022private,amid2022public} propose new private optimization methods under the setting where there are some public and non-sensitive data. 

The last direction is to relax the definition of DP and propose new DP-SGD variants. \citet{shi2021selective} tailor DP-SGD to SDP. Their method SDP-SGD first splits the text into the sensitive and non-sensitive parts, and applies normal SGD to the non-sensitive part while applying DP-SGD to the sensitive part respectively. Later, \citet{shi2022just} extend to large language models and propose a method, namely Just Fine-tune Twice to private fine-tuning with the guarantee of SDP. 

\subsection{Federated Learning Setting}
In the previous parts, we reviewed the related work on DP pre-trained models and DP fine-tuning models. Note that all the previous work only considers the central DP setting where all the training data samples are already collected before training, indicating that these methods cannot be applied to the federated learning (FL) setting. Compared to central DP, there are fewer studies on DP Federated Learning for NLP. \citet{mcmahan2017learning}  apply DP-SGD in the FedAvg algorithm to protect user-level privacy for LSTM and RNN architectures in the federated learning setting. Specifically, they first sample users with some probability, and then add Gaussian noise to model updates of the sampled users on the server side.  Based on this,  \citet{ramaswamy2020training} develop the first consumer-scale next-word prediction model.

Rather than adopting DP-SGD, \citet{kairouz2021practical} provides a new paradigm for DP-FL by using the Follow-The-Regularized-Leader (FTRL) algorithm, which achieves state-of-the-art performance, and it is recently improved by \citet{choquette2022multi,koloskova2023convergence,denisov2022improved,agarwal2021skellam}. 

It is notable that all the previous studies only consider shallow neural networks such as RNN and LSTM and do not consider the large language model. Until very recently, there have been some papers studying DP-FL fine-tuning. For example, \citet{wang2023can} consider the cross-device setting and use DP-FTRL to privately fine-tune. Moreover, they propose a distribution matching algorithm that leverages both private on-device LMs and public LLMs to select public records close to private data distribution. \citet{xu2023federated} deploy  DP-FL versions of Gboard Language Models \cite{hard2018federated} via DP-FTRL and quantile-based clip estimation method in \citet{andrew2021differentially}.  

\section{Embedding Vector Perturbation Based Methods}\label{sec:embedding}
Generally speaking, this type of approach considers privatizing the embedding vector for each token. Specifically, in this framework, the text data is first transformed into a vector (text representation) via some word embedding method such as Word2Vec \citep{mikolov2013efficient} and BERT. Then we use some DP mechanism to privatize each representation and train NLP models based on these privatized text representations. Due to the post-processing property of DP, we can see the main strength of this approach is any further training on these private embeddings also preserves the DP property, while gradient perturbation based methods heavily rely on the network structure. We can see that the main step of this method is to design the best private text representation. Note that since we need to privatize each embedding representation separately, the whole algorithm could be considered as an LDP algorithm, and thus, it can also be used in the LDP setting. It is also notable that different studies may consider different notions and levels of privacy.  In fact, most of the existing work considers the word level of privacy.

\subsection{Vanilla DP}
The most direct approach is to design private embedding mechanisms that satisfy the standard DP. \citet{lyu2020towards} first study this problem and they propose a framework. Specifically, firstly, for each word, the embedding module of such framework outputs a 1-dimensional real representation with length $r$, then it privatizes the vector via a variant of the Unary Encoding mechanism in  \citep{wang2017locally}. In order to remove the dependence of dimensionality in the Unary Encoding mechanism, they propose an Optimized Multiple Encoding, which embeds vectors with a certain fixed size. Their post-processing procedure was then improved by \citep{plant-etal-2021-cape}. In \citep{plant-etal-2021-cape}, it first gets the final layer representation of the pre-trained model for each token, then normalizes it with sequence and adds Laplacian noise, and finally trains this classifier with adversarial training. To further improve the fairness for the downstream tasks on private embedding, later \citet{lyu-etal-2020-differentially} propose to dropout perturbed embeddings to amplify privacy and a robust training algorithm that incorporates the noisy training representation in the training process to derive a robust target model, which also reduces model discrimination in most cases.

\citet{krishna-etal-2021-adept,habernal-2021-differential,DBLP:conf/sbp-brims/AlnasserB021} also study privatizing word embeddings. However, instead of using the Unary Encoding mechanism or dropout, \citet{krishna-etal-2021-adept,DBLP:conf/sbp-brims/AlnasserB021} propose ADePT, which is an auto-encoder-based DP algorithm. Let $\mathbf{u}$ be the input, an auto-encoder model consists of an encoder that returns a vector representation $\mathbf{r} = \text{Enc}(\mathbf{u})$ for the input $\mathbf{u}$, which is then passed into the decoder to construct an output $\mathbf{v} = \text{Dec}(\mathbf{r})$. In \citep{krishna-etal-2021-adept}, it first normalized the word embedded vector by some parameter $C$ {i.e.,} $w=\text{Enc(u)}\min\{1, \frac{C}{\|\text{Enc}(u)\|_2}\}$, then it adds Laplacian noise to the normalized vector $w$ and get  $\mathbf{r}$. Unfortunately, \citet{habernal-2021-differential} points out that ADePT is not differentially private by thorough theoretical proof. The problem of ADePT lies in the sensitivity calculation and could be remedied by adding calibrated noise or tighter bounded clipping norm. Later, \citet{Igamberdiev.2022.COLING} provides the source code of DP Auto-Encoder methods to improve reproducibility. Recently, \citet{maheshwari2022fair} proposed a method that combines differential privacy and adversarial training techniques to solve the privacy-fairness-accuracy trade-off in local DP. In their framework, first, the input text will be fed into encoders, then it will be normalized and privatized by using the Laplacian mechanism. Next, it will be fed into a normal classifier and adversarial training separately to combine a loss that contains normal classification loss and adversarial loss. They find that the model can improve privacy and fairness simultaneously. To further improve the performance, \cite{DBLP:journals/corr/abs-2309-10551} propose a Neighbourhood-Aware Differential Privacy (NADP) mechanism considering the neighborhood of a word in a pre-trained static word embedding space to determine the minimal amount of noise required to guarantee a specified privacy level. 

Besides the work on word-level privacy we mentioned above, recently, there have been some works studying sentence-level and token-level private embeddings. \citet{meehan2022sentence} propose a method, namely DeepCandidate, to achieve sentence-level privacy. They first put public and private sentences into a sentence encoder to get sentence embeddings. Then, they use a method, namely DeepCandidate, to choose the candidate sentence embeddings that are near to private embeddings. Finally, they use some DP mechanism to sample from the candidate embeddings for each private embedding. This method somehow solves the challenge of the sentence-level privacy problem by taking advantage of clustering in differential privacy. \cite{DBLP:journals/corr/abs-2309-06746} consider sentence-level privacy for private fine-tuning and propose DP-Forward fine-tuning, which perturbs the forward pass embeddings of every user’s (labeled) sequence. However, it is notable that they consider a variant of LDP called sequence local DP. \citet{DBLP:conf/acl/ChenMWCN0C23}  propose a novel Customized Text (CusText) sanitization mechanism that provides more advanced privacy protection at the token level.

\subsection{Metric DP}
In Metric DP for text data, each sample of the input can be represented as a string $x$ with at most $l$ words, thus, the data universe will be $W^\ell$ where $W$ is a dictionary. Also we assume that there is a word embedding model $\phi: W \mapsto \mathbb{R}^n$ and its associated distance $d(x, x')=\sum_{i=1}^l \|\phi(w_i)-\phi(w_i')\|_2$, where $x=w_1w_2\cdots w_l$ and $x'=w'_1w'_2\cdots w'_l$ are two samples. Thus, the goal is to design a mechanism for each $\phi(w_i)$ with the guarantee of Metric DP. Since we aim to randomize each $\phi(w_i)$ for each sample. The whole algorithm is also suitable for local metric DP with word-level privacy. 

\citet{feyisetan2020privacy} first study this problem. Generally speaking, their mechanism consists of two steps. The first step is perturbation, we add some noise $N$ to text vector $\phi(w_i)$ to ensure $\varepsilon$-LDP, where $N$ has the density probability function $p_N(z)\propto \exp(-\varepsilon \|z\|_2)$. The main issue of this approach is that after the perturbation, $\hat{\phi}_i$ may be inconsistent with the word embedding. That is, there may not exist a word $u$ such that $u=\hat{\phi}_i$. Thus, to address this issue, we need to project the perturbed vector into the embedding space. That is the second step. \citet{feyisetan2020privacy} show that the algorithm is $\varepsilon$-local Metric DP. 

Note that the method was later improved from different aspects. For example, \citet{xu2020mahala} reconsider the problem setting and they observe that the distance used in \citep{feyisetan2020privacy} is the Euclidean norm $d(x, x')=\sum_{i=1}^l \|\phi(w_i)-\phi(w_i')\|_2$, which cannot describe the similarity between two words in the embedding space. To address the issue, they propose to use the Mahalanobis Norm and modify the algorithm by using the Mahalanobis mechanism, which can improve performance. To further improve the utility in the projection step, \citet{xu2021utilitarian} further propose the Vickrey mechanism in case the first nearest neighbors are the original input or some rare words need large-scale noise to perturb and hard to find the corresponding words. In order to solve this problem, they use a hyperparameter in their algorithm to adjust the selection of the first and second nearest neighbors (words). To further allow a smaller range of nearby words to be considered than the multivariate Laplace mechanism, \cite{xu2021density,carvalho2021tem} propose an improved perturbation method via the Truncated Gumbel Noise. To further address the high dimensional issue, \citet{feyisetan2021private} uses the random projection for the original text representation to a lower dimensional space and then projects back to the original space after adding random noise to preserve DP. Besides, \citet{feyisetan2019leveraging} define the hyperbolic embeddings and use the Metropolis-Hastings (MH) algorithm to sample from hyperbolic distribution. However, it is remarkable that if we consider the LDP setting, then all the previous methods need to send real numbers to the server, which has a high communication cost. To address the issue, \citet{carvalho2021brr} proposes to use the binary randomized response mechanism by using binary embedding vectors. Recently, \citet{tang2020privacy} consider the case where different words may have different levels of privacy. They first divide the words into two types, and then add corresponding noise according to different levels of privacy. \citet{DBLP:conf/uai/ImolaKWAT22} recently proposed an optimal Meric DP mechanism for finite vocabulary, they then provided an algorithm that could quickly calculate the mechanism. Finally, they applied it to private word embedding.  Instead of developing new private mechanisms, there are also some studies on improving the embedding process. The previous metric DP mechanisms are expected to fall short of finding substitutes for words with ambiguous
meanings. To address these ambiguous words, \citet{DBLP:journals/corr/abs-2306-01457} provide a sense embedding and incorporate a sense disambiguation step prior to noise injection. \citet{DBLP:journals/corr/abs-2306-01471} account for the common semantic context issue that appeared in the previous private embedding mechanisms. They incorporate grammatical categories into the privatization step in the form of a constraint to the candidate selection and show that selecting a substitution with matching grammatical properties amplifies the performance in downstream tasks. \citet{qu2021natural} recently points out that \citep{lyu-etal-2020-differentially} does not address privacy issues in the training phase since the server needs users' raw data to fine-tune. Moreover, its method has a high computational cost due to the heavy encoder workload on the user side. Thus, \citet{qu2021natural} improve it and consider the federated setting where users send their privatized samples via some local metric DP mechanism to the server, and the server conducts privacy-constrained fine-tuning methods. Moreover, besides the text-to-text privatization given in \citep{feyisetan2020privacy} and the sequence private representation proposed by \citet{lyu-etal-2020-differentially}, \citet{qu2021natural} proposed new token-level privatization and text-to-text privatization methods. In the token representation privatization method, they add random noise using metric DP to token embedding and send it to the server. They add noise to the embedded token and output the closest neighbor token in the embedding space.

Instead of the local Metric DP, \citet{yue-etal-2021-differential} consider  UMLDP and propose SANTEXT and SANTEXT+ algorithms for text sanitization tasks. Specifically, they divide all the text into a sensitive token set $\mathcal{V}_S$ and a remaining token set $\mathcal{V}_N$. Then  $\mathcal{V}_S$ and $\mathcal{V}_N$ will use a privacy budget of $\epsilon$ and $\epsilon_0$ respectively via the composition theorem in LDP. After deriving token vectors, SANTEXT samples new tokens via local Metric DP with Euclidean distance. Compared with SANTEXT, SANTEXT+ samples new tokens when the original tokens are in sensitive set $\mathcal{V}_S$. They apply it to BERT pre-training and fine-tuning models.

While there are many studies on the benefits of private embedding with word-level privacy. There are also some shortcomings to such notion of privacy, as mentioned by \cite{mattern2022limits} recently. For example, in the previous private word embedding methods, we need to assume the length of the string for each sample is the same. Moreover, since we consider the word level of privacy, the total privacy budget will grow linearly with the length of the sample. To mitigate some shortcomings, \citet{mattern2022limits} propose an alternative text anonymization method based on fine-tuning large language models for paraphrasing. To ensure DP, they adopt the exponential mechanism to sample from the softmax distribution. They apply their method in fine-tuning models with GPT-2. 

Recently, \citet{10.1145/3543507.3583512} studied sentence-level private embedding in local metric DP. Borrowing the wisdom of normalizing sentence embedding for robustness, they impose a consistency constraint on their sanitization. They propose two instantiations from the Euclidean and angular
distances. The first one utilizes the Purkayastha mechanism \cite{DBLP:conf/ccs/WeggenmannK21}, and the other is upgraded from the generalized planar Laplace mechanism with post-processing. 

Very recently, besides pre-training and fine-tuning, private word embedding has also been used in the task of prompt tuning for Large Language Models. The goal of private prompt tuning is to protect the privacy of examples demonstrated in the prompt. Specifically, \citet{DBLP:journals/corr/abs-2305-06212} leverages the above private embedding methods to ensure local metric DP. To mitigate the performance degradation when imposing privacy protection, they propose a  privatized token reconstruction task motivated by the recent findings that the masked language modeling objective can learn separable deep representations. Then, the objective of privatized token reconstruction is to recover the original content of a privatized special token sequence from LLM representations. 

\section{Ensemble Model Based Methods} 
Unlike gradient perturbation and private embedding based methods, the general idea of ensemble model based methods is first we divide the whole private data into several subsets, then we {\bf non-privately} train a model for each private subset. To ensure privacy, for each time of inference or query, we will do a private aggregation for all models. Compared with the previous two types of approach, the main advantage of the ensemble model based method is the noise we add will be independent of the scale of the model or the dimension of the embedding space, indicating the noise is much smaller. However, the weakness is that here, we cannot release private embeddings or the private model, and each query or inference will cost a privacy budget. Generally speaking, based on different private aggregations, there are two types of approaches: the PATE-based method, and the Sample-and-Aggregation method. 

\subsection{PATE-based Method}

PATE \cite{papernot2016semi} was originally crafted for addressing classification tasks, and it incorporates both a private dataset and a public unlabeled dataset within its framework, drawing parallels to the principles of semi-supervised learning. PATE ensures DP by employing a teacher-student knowledge distillation framework consisting of multiple teacher models and a student model. In this setup, the student model acquires knowledge from the private dataset through knowledge distillation facilitated by the teacher models. The PATE framework consists of three key components: (i) Teacher Model Training: The private dataset is first shuffled and divided into $M$ distinct subsets. Each teacher model is subsequently trained on one of these subsets. (ii) Teacher Aggregation: To leverage the knowledge of the individual teacher models, their outputs are aggregated, and this aggregated information serves as supervision for the student model. Each of the trained teachers contributes their insights to guide the learning process of the student on the unlabeled public dataset. (iii) Student Model Training: The student model is trained on the public dataset using the guidance provided by the aggregated teacher models. This collaborative approach ensures that the student model learns from the unlabeled data while benefiting from the distilled knowledge of the teacher models.


In the context of classification tasks, a common practice involves leveraging the collective wisdom of teachers by using their noisy majority votes as labels to guide the students, thereby ensuring DP. However, when it comes to text generation tasks, the straightforward application of this framework encounters a significant challenge. This challenge arises because traditional text generation models generate words sequentially, typically from left to right. Consequently, a straightforward application of PATE to text generation necessitates the iterative unveiling of all teachers, word by word, which comes with substantial computational and privacy costs. To tackle this issue, an innovative solution was presented by \citet{DBLP:conf/nips/TianZHWZH22}, known as the SeqPATE framework. The SeqPATE framework initiates by generating pseudo-data using a pre-trained language model, simplifying the teachers' role to providing token-level guidance based on these pseudo inputs. In dealing with the inherent complexities of the expansive word output space and the accompanying noise, the framework introduces dynamic filtering of candidate words. This process focuses on selecting words with notably high probabilities. Additionally, the SeqPATE framework adopts a unique approach to aggregating teacher outputs. Instead of relying on voting, it involves an interpolation of their output distributions, offering a more refined and nuanced strategy for information fusion.


Recently, a notable development in the application of PATE, as reported by \citet{DBLP:journals/corr/abs-2305-15594}, extends its utility to the realm of private In-context learning, a domain where the primary objective revolves around safeguarding the privacy of downstream data embedded in discrete prompts. Departing from the conventional approach of training teacher models on distinct partitions of private data, this innovative method capitalizes on the private data to formulate distinct prompts for the Large Language Model (LLM). In the context of private knowledge transfer, the teachers take on the role of labeling public data sequences. Each teacher offers their perspective by voting on the most probable class labels for the private downstream task. On the student model front, a novel strategy is proposed, leveraging the data efficiency of the prompting technique. This approach entails using labeled public sequences to create new discrete prompts for the student model. The chosen prompt is subsequently deployed alongside the Large Language Model (LLM) to serve as the student model, effectively enhancing the overall efficiency and privacy of the In-context learning process.

\subsection{Sample-and-Aggregation-based Method}

In contrast to the PATE-based method, the Sample-and-Aggregation-based approach diverges significantly by omitting the presence of a public unlabeled dataset, rendering the incorporation of a student model unnecessary. Notably, the work by \citet{wu2023privacypreserving} delves into the realm of private In-context learning and provides a comprehensive protocol. The protocol encompasses the following crucial steps: The initial step involves the discreet partitioning of the dataset, specifically the private demonstration exemplars, into non-overlapping subsets of exemplars. Each of these subsets is then paired with relevant queries, culminating in the creation of exemplar-query pairs. For every exemplar-query pair, the Language Model's (LLM) API is invoked, eliciting a diverse set of responses. Subsequently, these individual responses generated by the LLM are aggregated in a manner compliant with differential privacy (DP) principles. The outcome is a privately aggregated model answer, which is then made available to the user. Furthermore, the study introduces two distinctive private aggregation schemes, thus enhancing the repertoire of options for preserving privacy in the context of In-context learning.


In a parallel exploration of private In-context learning, \citet{DBLP:journals/corr/abs-2309-11765} consider the scenarios involving an infinite number of queries. In lieu of generating private answers, their innovative approach revolves around the creation of synthetic few-shot demonstrations using the private dataset. This method involves augmenting each private subset with the information generated thus far, collectively contributing to the likelihood of generating the subsequent token. To mitigate the impact of noise prior to the private aggregation phase, the approach strategically curtails the vocabulary to include only tokens found within the top-K indices of the next-token probability. This is derived solely from the instructional content, entirely excluding any input from the private data. The probabilities associated with the next token generation, extracted from each individual subset, are then subjected to a private aggregation process, ensuring a nuanced and privacy-preserving amalgamation of information.

\section{Challenges and Future Directions}
\paragraph{DP  for LLMs.} Dealing with large-scale text data and training LLMs like GPT-4 are tough tasks in deep learning with DP. Due to the high dimensionality of embedding vectors, even adding small noise can have a significant influence on the training speed and performance of models. It is more severe for DP-SGD-based methods, which need high memory costs, and their per-example clipping procedure is time-consuming. These methods will be inefficient when they are applied to large language models. Thus, how to reduce the memory cost and accelerate the training or fine-tuning of DP-SGD become core concerns in gradient perturbation-based methods. Although there is some work in this direction, from Table \ref{overview} we can see most of the current studies are only for BERT, GPT-2, and T5, and there is still a gap in accuracy between private and non-private models and these methods still need catastrophic cost of memory compared with the non-private ones. Moreover, it is well known that we need a heavy workload on hyperparameter-tuning for large-scale models in the non-private case. From the privacy view, each try-on hyperparameter-tuning will cost an additional privacy budget, which makes our final private model cost a large privacy budget. Thus, how to efficiently and privately tune the hyperparameters in large models is challenging. 

Besides the central setting, from Table \ref{overview}, we can also see that DP training and In-context learning in the federated learning setting is still lacking in studies. Moreover, even for DP fine-tuning, we can see the current studies only focused on small models such as LaMDA, and there is still no study on private fine-tuning for LLMs in the federated learning setting. 

\paragraph{\bf Sentence-level Private Embedding} As we mentioned, in embedding vector perturbation-based methods, the core problem is how to derive a private embedding that can avoid information leakage while also having good performance for downstream tasks. These methods use variants of distances to extract the relationship between words in the embedding space and use different noises to obfuscate sensitive tokens. Besides, some work focuses on how to use these private embeddings in specific settings like the generation of synthetic private data, federated learning, and fine-tuning models. However, these papers only focus on word-level privacy and do not consider sentence-level privacy which is more practical in the NLP scenario. For example, even if we replace some sensitive words (like name) using private embedding methods in a question-answering system, we can still easily infer that person from some sentences. In total, we should not only consider the privacy issue of each word but also consider how to hide sentence structures and syntax in sentences. Thus, designing sentence-level private embeddings is an important but difficult problem in private language models. 

\paragraph{Private Inference.} It is notable that in this paper, we mainly discussed how to privately train and release a language model without leaking information about training data. However, in some scenarios (such as Machine Learning as a Service), we only want to use the model for inference instead of releasing the model. Thus, for these scenarios, we only need to perform inference tasks based on our trained model, while we do not want to leak information about training data. From the DP side, such private inference corresponds to the DP prediction algorithm, which is proposed by \cite{DBLP:conf/colt/DworkF18}. Compared with private training, DP inference for text data is still far from well-understood, and there are only few studies on it \cite{ginart2022submix,DBLP:journals/corr/abs-2205-13621}. 

\section*{Limitations}
First, in this paper, we mainly focused on the deep learning-based models for NLP tasks in the differential privacy model. Actually, there are also some studies on classical statistical models or approaches for NLP in DP, such as topic modeling \cite{DBLP:journals/corr/ParkFCW16a,DBLP:journals/tifs/ZhaoRYHZY21,DBLP:conf/emnlp/Huang021} and n-gram extraction \cite{DBLP:conf/nips/KimGKY21}. Secondly, due to the space limit, we did not discuss all the related work for DP-SGD, and we only focused on the work that uses DP-SGD to NLP-related tasks. Thirdly, while we tried our best to discuss all the existing work on deep learning-based methods for DP-NLP, we have to say that we may have missed some related work. Moreover, since we aim to classify all the current work into three categories based on their methods of adding randomness, there is still some work that does not belong to these three classes, such as \cite{bo-etal-2021-er,DBLP:conf/www/WeggenmannRAMK22}. To make our paper be consistent, we did not mention these works here. Fourthly, although DP can provide rigorous guarantees of privacy-preserving, it has also been shown that DP machine learning models can cause fairness issues. For example, they always have a disparate impact
on model accuracy \cite{DBLP:conf/nips/BagdasaryanPS19}. Finally, it is notable that in this paper, we did not discuss the narrow assumptions made by differential privacy, and the broadness of natural language and of privacy as a social norm. More details can be found in \cite{DBLP:conf/fat/BrownLMST22}.


\bibliography{anthology,custom}
\appendix

\section{Differential Privacy Preliminaries}\label{sec:dp}
Differential Privacy (DP) is a data post-processing technique, which guarantees data privacy by confusing the attacker. To be more specific, suppose there is one dataset noted as $S$, and we can get another dataset $S'$ by changing or deleting one data record in this dataset. Denote the output distribution 
when $S$ is the input as $P_1$, and the output distribution when $S'$ is the input as $P_2$, if $P_1$ and $P_2$ are almost the same, then we cannot distinguish these two distributions, i.e., we cannot infer whether the deleted or replaced data sample based on the output we observed.  The formal details are given by \citet{dwork2006calibrating}. Note that in the definition of DP, adjacency is a key notion. 
One of the commonly used adjacency definitions
is that two datasets $S$ and $S'$ are adjacent (denoted as $S\sim S'$) if $S'$
can be obtained by modifying one record in $S$.

\begin{definition}\label{def:dp}
Given a domain of dataset $\mathcal{X}$.
A randomized algorithm $\mathcal{A}: \mathcal{X}\mapsto \mathcal{R}$ is $(\varepsilon,\delta)$-differentially private (DP) if for all adjacent datasets $S,S'$ with each sample is in $\mathcal{X}$ and for all $T\subseteq \mathcal{R}$, the following holds
	$$\Pr(\mathcal{A}(S) \in T)\leq \exp(\varepsilon) \Pr(\mathcal{A}(S')\in T)+\delta.$$
	When $\delta=0$, we call the algorithm $\mathcal{A}$ is $\varepsilon$-DP. 
\end{definition}
\paragraph{Illustration:}
For example, let $\mathcal{X}$ be a collection of labeled product reviews, each belonging to a single individual, and let $\mathcal{R}$ be the parameters of a classifier trained on $\mathcal{X}$. If the classifier's training procedure $\mathcal{A}$ satisfies the DP definition above, an attacker's ability to find out whether a particular individual was present in the training data or not is limited by $\varepsilon$ and $\delta$.

In the definition of DP, there are two parameters $\epsilon$ and $\delta$. Specifically, $\epsilon$ measures the closeness between the output distribution when the input is $S$, and the output distribution when the input is $S'$, smaller $\epsilon$ indicates the two distributions are more indistinguishable, i.e., the algorithm $\mathcal{A}$ will be more private. In practice, we set $\epsilon=0.1-0.5$ as a high privacy regime. Informally, $\delta$ could be thought of as the probability ratio between the two distributions is not bounded by $e^\epsilon$. Thus, it is preferable to set $\delta$ as small as possible. In practice we always set $\delta$ as a value from $\frac{1}{n^{1.1}}$ to $\frac{1}{n^2}$, where $n$ is the number of samples in the dataset $S$. It is notable that besides $\epsilon$ and $(\epsilon, \delta)$-DP, there are also other definitions DP such as R\'{e}nyi DP \cite{DBLP:conf/csfw/Mironov17}, Concentrated DP \cite{DBLP:conf/tcc/BunS16,DBLP:journals/corr/DworkR16}, Gaussian DP \cite{https://doi.org/10.1111/rssb.12454} and Truncated CDP \cite{DBLP:conf/stoc/BunDRS18}. However, all of them can be transformed into the original definition of DP. Thus, in this survey, we mainly focus on Definition \ref{def:dp}. 

There are several important properties of DP, see \cite{TCS-042} for details. Here, we only introduce those which are commonly used in NLP tasks. The first one is post-processing, which means that any post-processing on the output of an $(\epsilon, \delta)$-DP algorithm will remain $(\epsilon, \delta)$-DP. Equivalently, if an algorithm is DP, then any side information available to the adversary cannot increase the risk of privacy leakage. 
\begin{proposition}
Let $\mathcal{A}:\mathcal{X}\mapsto \mathbb{R}$ be $(\epsilon, \delta)$-DP, and let $f: \mathcal{R}\mapsto \mathcal{R}'$ be a (randomized) algorithm. Then $f\circ \mathcal{A}:\mathcal{X}\mapsto \mathbb{R}'$ is $(\epsilon, \delta)$-DP. 
\end{proposition}

\paragraph{Example:}
Continuing with our scenario of training a review classifier under DP, let us imagine we take the model from the previous example, which was trained under $(\varepsilon, \delta)$-DP, and perform a domain adaptation by fine-tuning on a different dataset, this time without any privacy. The resulting model still remains $(\varepsilon, \delta)$-DP with respect to the original data, that is, privacy cannot be weakened by any post-processing.

The second property is the composition property. Generally speaking, the composition property guarantees that the composition of several DP mechanisms is still DP. 
\begin{proposition}[Basic Composition Theorem]\label{composition}
Let $\mathcal{A}_1, \mathcal{A}_2, \cdots, \mathcal{A}_k$ be $k$ sequence of randomized algorithms, where $\mathcal{A}_1: \mathcal{X}\mapsto \mathcal{R}_1$ and $\mathcal{A}_i: \mathcal{R}_1\times \cdots \mathcal{R}_{i-1}\times \mathcal{X} \mapsto \mathcal{R}_i$ for $i=2, \cdots, k$. Suppose that for each $i\in [k]$, $\mathcal{A}_i(a_1, \cdots, a_{i-1}, \cdot)$ is $(\epsilon_i, \delta_i)$-DP. Then the algorithm $\mathcal{A}: \mathcal{X}\mapsto \mathcal{R}_1\times \cdots \times \mathcal{R}_k$ that runs the algorithms $\mathcal{A}_i$ in sequence is $(\epsilon, \delta)$-DP with $\epsilon=\sum_{i=1}^k \epsilon_i$ and $\delta=\sum_{i=1}^k \delta_i$. 
\end{proposition}
The basic composition allows us to design complex algorithms by putting together smaller pieces. We can view the overall privacy parameter $\epsilon$ as a budget to be divided among these pieces. We will thus often refer to $(\epsilon, \delta)$ as the “privacy budget”: each algorithm we run leaks some information, and consumes some of our budget. Differential privacy allows us to view information leakage as a resource to be managed. For example, if we fix the privacy budget $(\epsilon, \delta)$, then making each $\mathcal{A}_i$ be $(\frac{\epsilon}{k}, \frac{\delta}{k})$-DP is sufficient to ensure the composition is $(\epsilon, \delta)$-DP. 

\paragraph{Example:} In most of the NLP tasks, we need to train a model by using variants of optimization methods, such as SGD or Adam. In general, these optimizers include several iterations to update the model, which could be thought of as a composition algorithm, and each iteration could be thought of as an algorithm. Thus, it is sufficient to design a DP algorithm for each iteration, and we can use the composition theorem to calculate the budget of the whole process. 

Besides the basic composition property, there are also several advanced composition theorems for $(\epsilon, \delta)$-DP, which could provide tighter privacy guarantees than the basic one. For example, consider each $\mathcal{A}_i, i\in[k]$ is $(\epsilon, \delta)$-DP. Then the basic composition theorem implies their composition is $(k\epsilon, k\delta)$-DP. However, this is not tight as we can use the advanced composition theorem to show their composition could be improved to $(O(\sqrt{k}\epsilon, O(k\delta))$-DP \cite{DBLP:conf/focs/DworkRV10}.  We refer to reference \cite{DBLP:conf/icml/KairouzOV15,DBLP:conf/tcc/MurtaghV16,DBLP:conf/ccs/0001M18} for details. 

The third property is the privacy amplification via subsampling. Intuitively, every differentially private algorithm has a much lower privacy parameter $\epsilon$ when it is run on a secret sample than when it is run on a sample whose identities are known to the attacker. And there, a secret sample can be obtained by subsampling as it introduces additional randomness. 
\begin{proposition}
Let $A$ be an $(\epsilon, \delta)$-DP algorithm. Now we construct the algorithm $B$ as follows: On input $D=\{x_1, \cdots, x_n\}$, first we construct a new sub-sampled dataset $D_S$ where each $x_i\in D_s$ with probability $q$.  Then we run algorithm $A$ on the dataset $D_S$. Then $B(D)=A(D_S)$ is $(\tilde{\epsilon}, \tilde{\delta})$-DP, where $\tilde{\epsilon}=\ln(1+(e^\epsilon-1)q)$ and $\tilde{\delta}=q\delta$. 
\end{proposition}
\paragraph{Example:} The subsampling property can be used for the private version of the stochastic optimization method. As in these methods, a common strategy is to use the subsampled gradient to estimate the whole gradient. 

It is notable that, besides subsampling, some other procedures could also amplify privacy, such as random check-in \cite{DBLP:conf/nips/BalleKMTT20}, mixing \cite{DBLP:conf/nips/BalleBGG19} and decentralization \cite{DBLP:conf/aistats/CyffersB22}. And for different subsampling methods, the privacy amplification guarantee is also different \cite{DBLP:journals/corr/abs-2105-10594,DBLP:conf/icml/ZhuW19,DBLP:conf/nips/BalleBG18}.

In the following, we will introduce some mechanisms commonly used in NLP tasks to achieve DP. 

We first give the definition of a (numeric) query.  The
query is simply something we want to learn from
the dataset. Formally, a query could be any function $f$
applied to a dataset $S$ and outputting a real valued vector, formally $f: \mathcal{X}\mapsto \mathbb{R}^d$. For example, numeric queries might return the sum of the gradient of the loss on all samples, number of females
in the database, or a textual summary
of medical records of all persons in the database
represented as a dense vector. Given a dataset $S$, a common paradigm for approximating $f(S)$ differentially privately is via adding some randomized noise. Laplacian noise and Gaussian noise are the most commonly used ones, which correspond to the Laplacian and Gaussian mechanisms, respectively. 
\begin{definition}[Laplacian Mechanism]\label{def:2}
Given a query $f : \mathcal{X} \mapsto\mathbb{R}^d$, the Laplacian Mechanism is defined as:
$\mathcal{M}_L(S,f,\epsilon)=q(S)+ (Y_1, Y_2, \cdots, Y_d),$
where $Y_i$ is i.i.d. drawn from a Laplacian Distribution $\text{Lap}(\frac{\Delta_1(f)}{\epsilon}),$ where $\Delta_1(f)$ is the $\ell_1$-sensitivity of the function $f$, {\em i.e.,}
$\Delta_1(f)=\sup_{S'\sim S'}||f(S)-f(S')||_1.$ For a parameter $\lambda$, the Laplacian distribution has the density function $\text{Lap}(\lambda) (x)=\frac{1}{2\lambda}\exp(-\frac{x}{\lambda})$. 
Laplacian Mechanism preserves $\epsilon$-DP.
\end{definition}
\begin{definition}[Gaussian Mechanism]
Given a query $f : \mathcal{X} \mapsto \mathbb{R}^d$, the Gaussian mechanism is defined as  $\mathcal{M}_F(S, f, \epsilon, \delta)= q(S)+\xi$ where $\xi\sim \mathcal{N}(0,\frac{2\Delta^2_2(f)\log(1.25/\delta)}{\epsilon^2}\mathbb{I}_d)$,  where $\Delta_2(f)$ is the $\ell_2$-sensitivity of the function $f$,
{\em i.e.,}
$\Delta_2(f)=\sup_{S\sim S'}||f(S)-f(S')||_2.$	Gaussian mechanism preserves $(\epsilon, \delta)$-DP when $0<\epsilon\leq 1$.
\end{definition}
From the previous two mechanisms, we can see that to privately release $f(S)$, it is sufficient to calculate the $\ell_1$-norm or $\ell_2$-norm sensitivity first and add random noise. Moreover, as $\Delta_2(f) \leq \Delta_1(f)$, the Gaussian mechanism will have lower error than the Laplacian mechanism, while we relax the definition from $\epsilon$-DP to $(\epsilon, \delta)$-DP. 

Instead of answering $f(S)$ privately, we also always meet the selection problem, i.e., we want to output the best candidate among several candidates based on some score of the dataset. The exponential mechanism is the one that can output a nearly best candidate privately. 
\begin{definition}[Exponential Mechanism]\label{def:4}
The Exponential Mechanism allows differentially private computation over arbitrary domains and range $\mathcal{R}$, parameterized by a score function $u(S,r)$ which maps a pair of input data set $S$ and candidate result $r\in \mathcal{R}$ to a real-valued score. With the score function $u$ and privacy budget $\epsilon$, the mechanism yields an output with exponential bias in favor of high-scoring outputs. 
Let $\mathcal{M}(S, u, \mathcal{R})$ denote the exponential mechanism, and $\Delta$ be the sensitivity of $u$ in the range $\mathcal{R}$, {\em i.e.,}
$\Delta=\max_{r\in \mathcal{R}}\max_{D\sim D'}|u(D,r)-u(D',r)|.$
Then if $\mathcal{M}(S, u, R)$ selects and outputs an element $r\in \mathcal{R}$ with probability proportional to $\exp(\frac{\epsilon u(S,r)}{2\Delta u})$, it  preserves $\epsilon$-DP.
\end{definition}
In the original definition of DP, we assume that data are managed by a trusted centralized entity that is responsible for collecting them and for deciding which differentially private data analysis to perform and to release. A classical use case for this model is the one of census data. Compared with the above model (which is called the central model), there is another model, namely the local DP model, where each individual manages his/her proper data and discloses them to a server through some differentially private mechanisms. The server collects the (now private) data of each individual and combines them into a resulting data analysis. A classical use case for this model is the one aiming at collecting statistics from user devices like in the case of Google’s Chrome browser. Formally, it is defined as follows.
\begin{definition}
For a data domain $\mathcal{X}$, a randomized algorithm  $\mathcal{A} :\mathcal{X}\mapsto \mathcal{R}$ is called $(\varepsilon, \delta)$-local DP (LDP) if for any $s, s' \in \mathcal{X}$ and $T \subseteq \mathcal{R}$ we have 
\begin{equation*}
   {\text{Pr}[\mathcal{A}(s)\in T] }\leq e^\varepsilon {\text{Pr}[\mathcal{A}(s')\in T] }+\delta.
\end{equation*}
\end{definition}
Compared with Definition \ref{def:dp}, we can see that here the main difference is the inequality holds for all elements $s, s'\in \mathcal{X}$ instead of all adjacent pairs of the dataset. In this case, each individual could ensure that their own disclosures are DP via the randomizer $\mathcal{A}$. In some sense, the trust barrier is moved closer to the user. While this has the benefit of providing a stronger privacy guarantee, it also comes at a cost in terms of accuracy.


It is notable that besides the central DP and local DP model, there are also other intermediate models such as shuffle model \cite{DBLP:conf/eurocrypt/CheuSUZZ19} and multi-party setting \cite{DBLP:conf/nips/PathakRR10}. However, as they are seldom studied in NLP, we will not cover these protocols in this survey. 

\end{document}